\documentclass[letterpaper, 10 pt, journal, twoside]{IEEEtran}

\usepackage[left=48pt,top=54pt,right=48pt,bottom=45pt]{geometry}

\usepackage{textcomp}
\usepackage[utf8]{inputenc}
\usepackage{moreverb,url}
\usepackage{times}
\usepackage{multicol}
\usepackage{amsmath,amsfonts,amssymb,mathtools,leftidx}
\usepackage{wrapfig}
\usepackage{float}
\usepackage{xcolor}
\usepackage{graphicx}
\usepackage{multirow}
\usepackage{mathrsfs}
\usepackage{nccmath}
\usepackage{balance}
\usepackage{afterpage}
\newtheorem{remark}{Remark}

\usepackage{tabularx}
\usepackage[noadjust]{cite}
\usepackage[colorlinks,bookmarksopen,bookmarksnumbered,citecolor=red,urlcolor=red]{hyperref}
\usepackage{booktabs}
\usepackage[ruled,vlined,linesnumbered,algo2e]{algorithm2e} 

\usepackage{algorithm,algpseudocode}

\makeatletter
\newcommand\fs@spaceruled{\def\@fs@cfont{\bfseries}\let\@fs@capt\floatc@ruled
    \def\@fs@pre{\vspace{5\baselineskip}\hrule height.8pt depth0pt \kern2pt}%
    \def\@fs@post{\kern2pt\hrule\relax}%
    \def\@fs@mid{\kern2pt\hrule\kern2pt}%
    \let\@fs@iftopcapt\iftrue}
\makeatother

\IEEEoverridecommandlockouts

\begin{document}

\title{ACD-EDMD: Analytical Construction for Dictionaries of Lifting Functions in Koopman Operator-based Nonlinear Robotic Systems}

\author{Lu Shi and Konstantinos Karydis

\thanks{Manuscript received: September 8, 2021; Accepted November 11, 2021.}
\thanks{This paper was recommended for publication by Editor Dr. Lucia Pallottino upon evaluation of the Associate Editor and Reviewers' comments.} %
\thanks{We gratefully acknowledge the support of NSF under grants \# IIS-1910087, \# CMMI-2046270 and \# CMMI-2133084, and of ONR under grant \# N00014-19-1-2264. Any opinions, findings, and conclusions or recommendations expressed in this material are those of the authors and do not necessarily reflect the views of the funding agencies.}
\thanks{The authors are with the Dept. of Electrical and Computer Engineering, University of California, Riverside. 
	Email: {\{lshi024, karydis\}@ucr.edu}.}
}

\markboth{IEEE Robotics and Automation Letters. Preprint Version. Accepted November, 2021}%
{Shi and Karydis: Analytical Construction of Dictionaries for Koopman Operator-based Systems}

\maketitle

\begin{abstract}
    Koopman operator theory has been gaining momentum for model extraction, planning, and control of data-driven robotic systems. The Koopman operator's ability to extract dynamics from data depends heavily on the selection of an appropriate dictionary of lifting functions. In this paper we propose ACD-EDMD, a new method for Analytical Construction of Dictionaries of appropriate lifting functions for a range of data-driven Koopman operator based nonlinear robotic systems. The key insight of this work is that information about fundamental topological spaces of the nonlinear system (such as its configuration space and workspace) can be exploited to steer the construction of Hermite polynomial-based lifting functions. We show that the proposed method leads to dictionaries that are simple to implement while enjoying provable completeness and convergence guarantees when observables are weighted bounded. We evaluate ACD-EDMD using a range of diverse nonlinear robotic systems in both simulated and physical hardware experimentation (a wheeled mobile robot, a two-revolute-joint robotic arm, and a soft robotic leg). Results reveal that our method leads to dictionaries that enable high-accuracy prediction and that can generalize to diverse validation sets. The associated GitHub repository of our algorithm can be accessed at \url{https://github.com/UCR-Robotics/ACD-EDMD}.
\end{abstract}


\section{Introduction}
\IEEEPARstart{K}{oopman} operator theory and associated numerical methods have been widely applied for system identification, state estimation, and control (e.g.,~\cite{kaiser2020KoopApp,korda2018KoopApp,abraham2019KoopApp,huang2020KoopmanApp}).
In an effort to handle approximate models (or lack thereof) that serve as a target for motion control of (nonlinear) robotic systems, methods based on Koopman operator theory are increasingly used in the context of robotics. Recent examples include 
modeling and control of a tail-actuated robotic fish~\cite{mamakoukas2021Taylor}, trajectory control of micro-aerial vehicles~\cite{shi2020ccta}, dynamics estimation for a spherical robot~\cite{abraham2017RKexample}, model extraction for a simulated lunar lander system~\cite{broad2020RKexample}, as well as model extraction and control for soft robotic arms~\cite{bruder2019koopsoft,bruder2019koopsoftD,bruder2020koopsoft,haggerty2020koopsoft} and underwater soft robots~\cite{castano2020koopsoft}.

A critical challenge inherent to all methods employing Koopman operator theory is the choice of a proper set of lifting functions (typically called the dictionary). The lifting functions are crucial because they serve as the basis to construct an infinite-dimensional linear approximation of a (nonlinear) system's state evolution. Poor choice of the lifting function can significantly impact the estimation accuracy of the Koopman operator and the higher-dimensional linearized dynamics. This paper presents a new method to analytically construct dictionaries of lifting functions for Koopman operator based data-driven nonlinear robotic systems.

Existing works regarding construction of dictionaries for Koopman operator based systems fall under four main directions. 1) The first one is empirically~\cite{williams2015EDMD}. For example, Legendre polynomials can make the observation matrix be block diagonal, Hermite polynomials are best suited to problems where data are normally distributed, and radial basis functions are effective for systems with complex geometry~\cite{boyd2001polynomials,wendland1999RBF}. However, empirical approaches can be time- and effort-demanding and cannot guarantee generalization to and efficiency in new cases. 2) Another direction is to rely on machine learning to derive the dictionary~\cite{li2017ML,yeung2019learning,takeishi2017learning}. While such methods have stronger generalization capacity, they require significant tuning (e.g., in the case of neural networks the number of layers, number of units per layer, etc.), and large amounts of training data.  The latter can in practice pose a significant challenge in robotics applications where data are in principle small. 3) A third direction is when the original model has a special structure or some underlying fundamental model is known (or assumed). For example, elementary functions can be used to map a system to an equivalent polynomial form via Lie derivatives~\cite{netto2020LieD}. However, algorithms of that nature require the original system to be a linear combination of elementary functions. Another instance is to analyze the geometric relation between a system model and its Koopman operator to obtain a dictionary~\cite{bollt2019geometric}. Although model-based methods of that spirit can be useful in controller design and dimension reduction, they cannot offer model identification. The latter is useful as uncertainty can affect robot behavior to the extent that an otherwise well-tuned model is no longer representative of the underlying interaction dynamics between the robot and its underlying environment~\cite{karydis2015IJRR,karydis2016ISER}.
4) A fourth approach is to compare and optimize over multiple sets of lifting functions to find a proper basis set~\cite{haggerty2020koopsoft}, but still the question of how to select the sets to begin with in an efficient manner remains.

Different from existing related methods, our proposed approach exploits the fact that robotic systems have certain characteristic properties that can be acquired without knowledge of their exact dynamical models. Properties considered herein are the system's configuration space, and  
its workspace. 
These properties reveal fundamental information about system states and dynamics, and can provide intuition on how to select lifting functions required for Koopman operator approximation. 

In this paper we propose ACD-EDMD, a general and analytical methodology to formalize the construction of lifting functions based on such system characteristic properties. We show how fundamental topological spaces and Cartesian products thereof can be mapped to a basis of Hermite polynomials and Kronecker products thereof which serve as the dictionary of lifting functions. 
We further show that the resulting dictionary is complete and leads to an estimated Koopman operator with provable guarantees of convergence to the true one, in the limit and provided that the observables are weighted bounded. At the same time the resulting dictionary is simple to implement. 

We evaluate the efficacy of our proposed method using a series of simulated and hardware experiments. We consider a differential drive robot in both simulation and physical experimentation, and a rigid robotic arm comprising two revolute joints in simulation. In these cases we use the configuration space of the robots. The method can also apply to soft robots (whereby their configuration space is ill-defined), by considering their workspace instead. Physical experimentation using a soft robotic leg that can bend and extend confirms that our approach can apply uniformly across rigid and soft robots, and further demonstrates its practical utility. Results obtained by ACD-EDMD are also compared against other nonlinear dynamics identification approaches in terms of prediction accuracy and training time, to demonstrate our method's efficacy. 


\section{Preliminary Technical Background}\label{sec:preliminary}
\subsection{Koopman Operator Theory and Extended Dynamic Mode Decomposition (EDMD)}
In Koopman operator theory~\cite{koopman1931koopman}, the infinite-dimensional linear operator governs the evolution of observables $g(x_t)$; the nonlinear evolving operator $f$ of the original system is represented by Koopman modes, eigenvalues and eigenfunctions. We use Koopman operator theory and EDMD to extract the nonlinear system dynamics from data. The relevant background is introduced in this section.

Consider the nonlinear system $x_{t+1} = f(x_t)$, 
where $ x\in \mathbb{R}^{n_x}$. 
The propagation law of observables $g$ with the Koopman operator $\mathcal{K}:\mathcal{F} \to \mathcal{F}$ is
$\mathcal{K} g(x_t)= g(f(x_t)).$
Then, decomposing the full state observable~\cite{williams2015EDMD} $g(x)=x$ with $N$ Koopman modes $v_n$, eigenvalues $\lambda _n$ and eigenfunctions $\varphi_n$, we obtain
\begin{equation}\label{eq:extendPredicion}
\begin{medsize}
    x_{t+1} = g(f(x_t))= \mathcal{K} g(x_t) \to x_{t+1}  =\sum_{n=1}^N v_n\lambda_n\varphi_n(x_t)\enspace . 
\end{medsize}
\end{equation}

Given snapshots of state measurements {\small $X = [x_1,x_2,\dots,x_{M},x_{M+1}]$}, the Koopman operator $\mathcal{K}$ can be approximated from the observations via EDMD, which generates a finite dimensional approximation $K:\mathcal{F}_N \to \mathcal{F}_N$ of the Koopman operator $\mathcal{K}:\mathcal{F} \to \mathcal{F}$. It employs a dictionary of functions to lift state variables to a space where observable dynamics is approximately linear. 

One of the most critical steps is to choose a proper dictionary for lifting the original states, $\mathcal{D} = span \left\{\psi_{1},\psi_{2}, \ldots, \psi_{N}\right\}$. 
If we set the vector-valued dictionary as $\mathbf{\Psi}(x_m)=[\psi_{1}(x_m), \ldots, \psi_{N}(x_m)]$, the Koopman operator can be approximated by minimizing the total residual between snapshots, i.e. 
\begin{equation}\label{eq:Minimization}
    J =\frac{1}{2} \sum_{m=1}^{M}\left(\mathbf{\Psi}\left(x_{m+1}\right)-\mathbf{\Psi}\left(x_{m}\right) K\right)^{2}.
\end{equation}

The least-squares problem can be solved by truncated singular value decomposition, yielding
    \begin{equation}\label{eq:estimation}
        K \triangleq \boldsymbol{G}^{\dagger} \boldsymbol{A}, \enspace
        \text{where }
        \begin{cases}
      \boldsymbol{G}=\frac{1}{M} \sum_{m=1}^{M}  \mathbf{\Psi}_m^{*} \mathbf{\Psi}_m\enspace,\\
      \boldsymbol{A}=\frac{1}{M} \sum_{m=1}^{M}  \mathbf{\Psi}_m^{*}  \mathbf{\Psi}_{m+1}\enspace,
    \end{cases}
    \end{equation}
with $^\dagger$ denoting the pseudoinverse, and $T$ and $^*$ denoting transpose and conjugate transpose operations, respectively.

With $K$ via~\eqref{eq:estimation}, we obtain 
\begin{equation}\label{eq:KoopmanDecomposition}
    \begin{cases}
        v_n = (w_n^*B)^T \enspace,\\
        \lambda_n \xi_n = K\xi_n \enspace,\\
        \varphi_n = \mathbf{\Psi}_t\xi_n \enspace,
    \end{cases}
\end{equation}
where $\xi_n$ is the $n$-th eigenvector, $w_n$ is the $n$-th left eigenvector of $K$ scaled so $w_n^T\xi_n = 1$, and $B$ is the matrix of appropriate weighting vectors so that $x=(\mathbf{\Psi} B)^T$~\cite{williams2015EDMD}. Now we can describe the evolution of the original nonlinear system using the estimated Koopman operator by plugging expressions~\eqref{eq:KoopmanDecomposition} back to~\eqref{eq:extendPredicion}. Control inputs can be readily incorporated to the definition of $\Psi$ as an augmented state~\cite{proctor2018EDMDc}.


\subsection{EDMD with Dictionary Learning (EDMD-DL)}

For high-dimensional and highly nonlinear systems, machine learning methods can help make selections on lifting functions~\cite{li2017EDMDLearning}. In this method, the lifting functions $\mathbf{\Psi}$ are trained and represented by an artificial neural network. Thus, EDMD is coupled with the trainable dictionary. 

In EDMD-DL~\cite{li2017EDMDLearning}, given data measurements $X$ the dictionary vector $\mathbf{\Psi}(x_m)$ is parameterized by a universal function approximator, i.e. $\Psi(x) = \Psi(x;\theta)$ for $\theta \in \Theta$ to be varied. Then, a feedforward 3-layer neural network is designed to approximate $\Psi$ that solves the minimization problem~\eqref{eq:Minimization} as
$$\Psi(x) = W_{out}h_3+b_{out}$$
$$h_{k+1} = tanh(W_k h_k+b_k),\enspace k = 0,1,2\enspace.$$

The set of all trainable parameters  is $\theta = \{W_{out},b_{out},\{W_k,b_k\}_{k=0}^2\}$. By iterating over two steps: (1) fix $\theta$, optimize $K$ as a least-square problem; then (2) fix
$K$, optimize $\theta$ as a standard machine learning problem, the dynamics can be estimated until convergence.


\subsection{Sparse Identification of Nonlinear Dynamical Systems (SINDy)}
SINDy~\cite{brunton2016SINDy} has been proposed for extracting governing equations of nonlinear systems from data. The method considers that only a few important terms govern the dynamics of an underlying model and uses sparse regression to determine the fewest terms required to accurately illustrate the system's state evolution based on observed (time-varying) data series. 

Letting the system be $\dot{x}_t = f(x_t)$, where $f(x_t)$ represents the dynamic constraints that describe propagation rules and is to be identified by data, snapshots of states $x_t$ and their derivatives $\dot{x}_t$ are collected or estimated and arranged into two data matrices $X$, $\dot{X}$. Then, a library $\Theta(X)$ consisting of candidate nonlinear functions of the column of $X$ is designed. Entries in this matrix of nonlinearities can be chosen with significant freedom. One example consists of constant, polynomial and trigonometric terms~\cite{brunton2016SINDy}, i.e.
\begin{equation}\label{eq:SINDy fcn}
    \Theta(X) = \left[ \begin{matrix}
    1 & X & X^2 & \sin(X) & \cos(X)
    \end{matrix}\right]
\end{equation}
A sparse regression problem to calculate the sparse vectors of coefficients $\Xi = [\xi_1,\xi_1,\dots,\xi_{n_x}]$ is setup as $\dot{X} = \Theta(X)\Xi$. Once $\Xi$ has been determined, the system can be approximated as 
$\dot{x} = f(x) = \Xi^T(\Theta(x^T))^T$. 
The process can also be extended to include inputs~\cite{brunton2016SINDyc}.

\subsection{Convergence of the EDMD operator}\label{sec:convergence}
It has been shown that as $M\to \infty$ the operator $K_{N,M}$ converges to $K_{N}$, the orthogonal projection of $\mathcal{K}$ on the subspace spanned by the lifting functions~\cite{klus2016convergence}. 
That result has been extended to analyze the convergence of $K_{N}$ to the actual Koopman operator $\mathcal{K}$~\cite{korda2018convergence}. In Remark~\ref{rmk:convergence} below we list an important result from~\cite{korda2018convergence}, which we build upon in our own theoretical analysis presented next in Section~\ref{sec:approach}.  

\begin{remark}\label{rmk:convergence}
(Adapted from~\cite{korda2018convergence}) \textit{If 1) the Koopman operator $\mathcal{K}$ is bounded; 2) the lifting functions $\psi_1, \dots, \psi_N$ are selected from an orthonormal basis of $\mathcal{F}$, then the sequence of operators $\mathcal{K}_{N}$ converges to $\mathcal{K}$ as $N \to \infty$.}
\end{remark}

Thus, if the chosen dictionary of lifting functions satisfies the above two conditions, the EDMD-estimated operator converges to the actual one.

\subsection{Property of Hermite Polynomials}\label{sec:HP}
 
Hermite functions serve as an orthonormal basis (complete orthonorml set) for the Hilbert space~\cite{celeghini2021hermite,reed1972hermite}, which is useful in our theoretical analysis. Remark~\ref{rmk:Hermite} below elaborates. 

\begin{remark}\label{rmk:Hermite}
Hermite polynomials form an orthogonal basis of the Hilbert space of functions $g(x)$ satisfying $\int_{-\infty}^{\infty}|g(x)|^{2} w(x) d x<\infty $, in which the inner product is given by the integral $\langle g_1, g_2\rangle=\int_{-\infty}^{\infty} g_1(x) \overline{g_2(x)} w(x) d x$, including the Gaussian weight function $w(x)$.
\end{remark}

\section{Dictionary Construction}\label{sec:approach}
In this section we present our main technical result. The key insight is that fundamental topological spaces and Cartesian products thereof can be mapped to a basis of Hermite polynomials and Kronecker products thereof. The latter produces the dictionary of lifting functions, which, as we show in the following, enjoys provable convergence guarantees of estimated Koopman operator to the true one when the observables are weighted bounded. Fundamental topological spaces in the context of robotics include the robot's configuration space and its workspace. 

\begin{figure}[!t]
\vspace{0pt}
\centering
\includegraphics[clip,width = 0.45\textwidth]{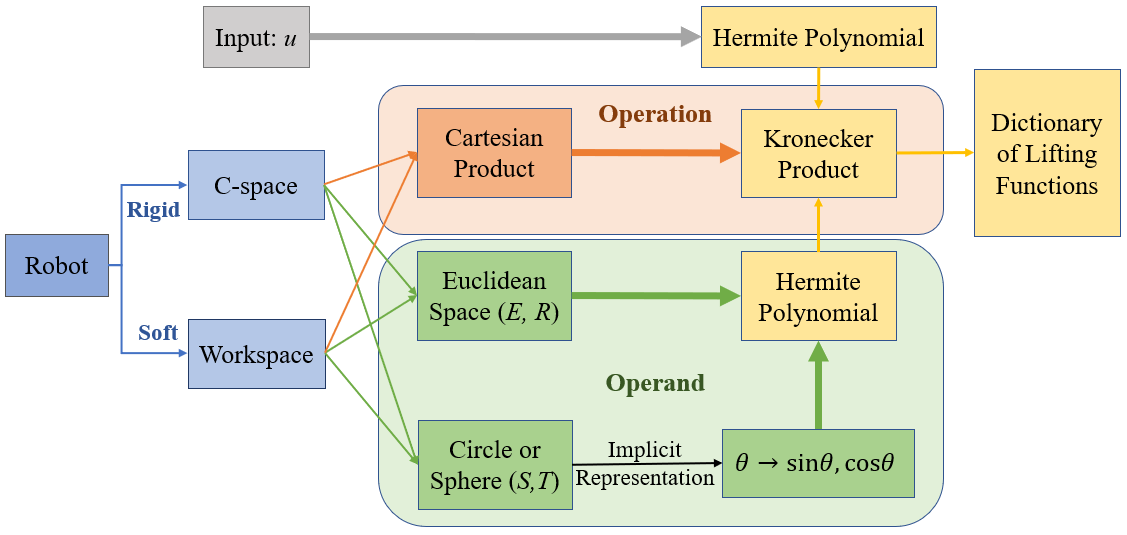}
\vspace{-6pt}
\caption{Illustration of our proposed method pipeline. Thin arrows indicate the computing sequence and the thick arrows represent how to map fundamental topological spaces to lifting functions. 
Hermite polynomials establish lifting functions for lower-dimensional spaces that work as the `operand.' The dictionary of higher-dimensional spaces is formed as the Kronecker product (the `operation') of sets of Hermite polynomials.}
\label{fig:algorithm}
\vspace{-15pt}
\end{figure}

Importantly, in our approach we consider both rigid and soft robots. This serves two key purposes. 1) To demonstrate that employing a robot's fundamental topology to yield lifting functions is general and can apply across robotic embodiments. 2) To offer a baseline in which the configuration space or the workspace topology might be preferred one over another. 
For rigid robots we consider the configuration space. However, soft robots are often considered to contain an infinite number of degrees of freedom, hence the problem of constructing their configuration space is ill-defined~\cite{jing2017overview}. Thus, we consider the workspace instead for the case of soft robots. 
A flowchart of the proposed method is given in Fig.~\ref{fig:algorithm}.

\begin{table*}[ht]
\vspace{0pt}
    \caption{Configuration Space Topology and Associated Lifting Functions}\label{tab:configuration_space}
    \vspace{-6pt}
    \centering
    \begin{tabular}{c c c c c}
    \toprule
        \multirow{2}{*}{System} & \multirow{2}{*}{Topology} & Sample & $\#$ of &Lifting Function $D(x)$ \\
        & &Representation& states & (recall notation: $H_1(x)=[H^0(x),H^1(x)]$)\\
        \midrule
        Point on a Line & $\mathbb{E}^1$ or $\mathbb{R}^1$& $x$ & 1 &$H_1(x)$ \\
        \midrule
        Point on a Plane & $\mathbb{E}^2$ or $\mathbb{R}^2$ & $(x,y)$ & 2 &$kron(H_1(x),H_1(y))$ \\
        \midrule
        Point on a 3D Space & $\mathbb{E}^3$ or $\mathbb{R}^3$ & $(x,y,z)$ & 3 &$kron(kron(H_1(x),H_1(y)),H_1(z))$ \\
        \midrule
        Spherical Pendulum & $\mathbb{S}^2$ & $ (\theta,\phi)$ & 2 &$kron\left(kron(H_1(\sin(\theta)),H_1(\cos(\theta))),kron(H_1(\sin(\phi)),H_1(\cos(\phi)))\right)$\\
        \midrule
        2R Robot Arm & $\mathbb{T}^2 = \mathbb{S}^1\times \mathbb{S}^1$ & $ (\theta,\phi)$ & 2 &$kron\left(kron(H_1(\sin(\theta)),H_1(\cos(\theta))),kron(H_1(\sin(\phi)),H_1(\cos(\phi)))\right)$ \\
        \midrule
        Rotating Sliding Knob & $\mathbb{E}^1\times \mathbb{S}^1$ & $(x,\theta)$ & 2 & $kron(H_1(x),kron(H_1(\sin(\theta)),H_1(\cos(\theta))))$ \\
        \midrule
        Wheeled Robot & $\mathbb{R}^2\times \mathbb{S}^1$ & $ (x,y, \theta)$ & 3 & $kron(kron(H_1(x),H_1(y)),kron(H_1(\sin(\theta)),H_1(\cos(\theta))))$ \\
        \bottomrule
    \end{tabular}
    \vspace{-16pt}
\end{table*}

\subsection{Analytical Construction of Lifting Functions for Fundamental Topological Spaces}

The Euclidean space $\mathbb{E}^n$ is very frequently used in the context of robotics. We use the Hermite polynomials to construct the basis functions.\footnote{There are two different definitions of Hermite polynomials~\cite{maheshwari2014Hermite}, the ``probabilist's" and the ``physicist's." We consider the ``probabilist's" Hermite polynomial; yet, the same analysis applies to the ``physicist's" version too.}  The Hermite polynomials form an orthogonal basis of the Euclidean space (or the Hilbert for higher dimensions). 
Non-Euclidean spaces which are often employed include the circle $S^1$, the sphere $S^n$, and the torus $T^n$. 
We elect to represent non-Euclidean spaces by an implicit parameterization of unit complex numbers so that the explicit variables are mapped to a higher dimensional space, e.g., an angle $\theta \in S^1$ will be mapped to $[\sin(\theta), \cos(\theta)] \in \mathbb{E}^2$. Doing so enables use of Hermite polynomials for a range of both Euclidean and non-Euclidean spaces.

Higher-dimensional spaces can be expressed as the Cartesian product of lower-dimensional spaces that contains the union of these spaces. 
For topological spaces constructed as the Cartesian product, we compute the Kronecker product of the lifting functions of the lower-dimensional spaces as the dictionary for the higher-dimensional space. The Kronecker product can be viewed as a form of vectorization (or flattening) of the outer product so it contains the results of all the elements multiplied from the two sets. 

We first construct the lifting functions for the state, $D(x)$; some examples of lifting functions are listed in Table~\ref{tab:configuration_space}. Note that we have considered zero- and first-order Hermite polynomials (denoted by $H^0$ and $H^1$, respectively) but ACD-EDMD may apply when considering higher-order terms as well; investigating this direction is part of future work. For clarity, we set $H_1(x)=[H^0(x),H^1(x)]$. 
For the control input, $D(u)$ is computed by the zero- and first-order Hermite polynomials and Kronecker products thereof similarly to the $\mathbb{R}^n$ cases for states as in Table~\ref{tab:configuration_space}. \footnote{This way to include the control input in the dictionary is consistent with that of other related methods that append the control input to the state~\cite{kaiser2020KoopApp}.} 
Having computed $D(x)$ and $D(u)$, the complete dictionary is then formed as the Kronecker product between lifting functions for states and inputs, i.e. $\mathcal{D} = kron(D(x), D(u))$ (see Fig.~\ref{fig:algorithm}).
\vspace{-3pt}


\subsection{Theoretical Analysis of ACD-EDMD}

The lifting functions comprising dictionary $D$ are Kronecker products of Hermite polynomials in a single dimension. This set of basis functions is \textbf{simple to implement}, and conceptually related to approximating the
Koopman eigenfunctions with a Taylor expansion~\cite{williams2015EDMD}. Furthermore, because they are orthogonal with respect to Gaussian weights, the matrix $G$ in~\eqref{eq:estimation} will be diagonal if the data are drawn from a normal distribution, which can be beneficial numerically.

The Hermite polynomials also form a \textbf{complete} orthogonal basis of the Hilbert space of the weights with exponential decay, i.e. the linear span of the basis is dense~\cite{maheshwari2014Hermite}. We can approximate the Koopman operator $\mathcal{K}$ with arbitrarily high accuracy by using sufficiently large number of terms of the basis functions when the observables are weighted bounded in the Euclidean (Hilbert) space. In other words, the EDMD operator using the proposed dictionary enjoys provable conditioned \textbf{convergence} guarantees. Based on the technical preliminaries discussed in Sections~\ref{sec:convergence} and~\ref{sec:HP}, we can deduce the following theorem.

\textbf{Theorem 3.1} \textit{If the \textbf{observable functions} $g \in \mathcal{F}$ satisfy}
\begin{equation}\label{eq:thm_cnd}
    \int_{-\infty}^{\infty}|g(x)|^{2} w(x) d x<\infty 
\end{equation}
\textit{with the inner product defined as}
$
\langle g_1,g_2\rangle=\int_{-\infty}^{\infty}g_1(x)\overline{g_2(x)}w(x)d x \enspace,
$
\textit{where $\overline{g}$ denotes the conjugate function, and weighting function $w(x)$ is the Gaussian weight function, then the operator $K_{N,M}$ estimated by ACD-EDMD converges to $\mathcal{K}$ as the number of samples $M$ and number of used lifting functions $N$ go to infinity.}

\textit{Proof:} The proof is decomposed into four steps. 1) The convergence of $K_{N,M}$ to $K_{N}$ is proven in~\cite{klus2016convergence} as $M\to \infty$. 
2) Condition~\eqref{eq:thm_cnd} dictates that observables $g(x)$ are bounded. This implies that the Koopman operator $\mathcal{K}$ calculated by these observables is also bounded and hence satisfies the first condition of Remark~\ref{rmk:convergence}. 
3) Per Remark~\ref{rmk:Hermite}, the Hermite polynomials form an orthogonal (which implies orthonormal) basis of the Hilbert space for all the weighted bounded observables $g$, and hence the second condition of Remark~\ref{rmk:convergence} is also satisfied. 4) We have established so far that individual Hermite polynomials satisfy the two conditions of Remark~\ref{rmk:convergence}. Here we consider the Kronecker product of Hermite polynomials, which nonetheless does not affect the aforementioned properties. Thus, with the proposed lifting functions and if the observables are well-designed (weighted bounded), we have $K_{N,M} \longrightarrow \mathcal{K}$ as $M\longrightarrow\infty$ and $N\longrightarrow\infty$.  \hfill$\blacksquare$

\section{Experimental Evaluation}
We evaluate the efficacy of our proposed analytical method to generate lifting functions for Koopman operator based nonlinear robotic systems using a series of both simulated and hardware experiments. We consider a differential drive robot (ROSbot 2.0) in both simulation and physical experimentation, a rigid robotic arm comprising two revolute joints in simulation, and a soft robotic leg~\cite{liu2020SoRX} that can bend and extend in physical experimentation. Results from ACD-EDMD are compared against those attained by other related methods introduced in Section~\ref{sec:preliminary}: 1) Classical EDMD with dictionary that contains the direct sum of Hermite polynomials of up to second-order terms in all states; 2) EDMD-DL with 25 dictionary outputs (includes one constant non-trainable map, the coordinate projection non-trainable maps and the rest trainable lifting functions); 3) SINDy with nonlinear functions introduced in~\eqref{eq:SINDy fcn} solved with least absolute shrinkage and selection operator (LASSO)~\cite{tibshirani1996LASSO}.

\subsection{Differential Drive Robot - Simulation}

\begin{figure}[!t]
\vspace{0pt}
\centering
\includegraphics[clip,width = 0.14\textwidth]{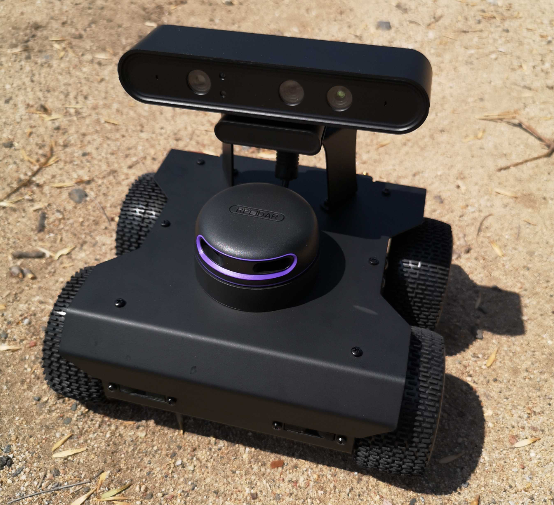}
\hspace{3pt}
\includegraphics[trim={0.4cm 0.25cm 0.58cm 0.3cm},clip,width = 0.14\textwidth]{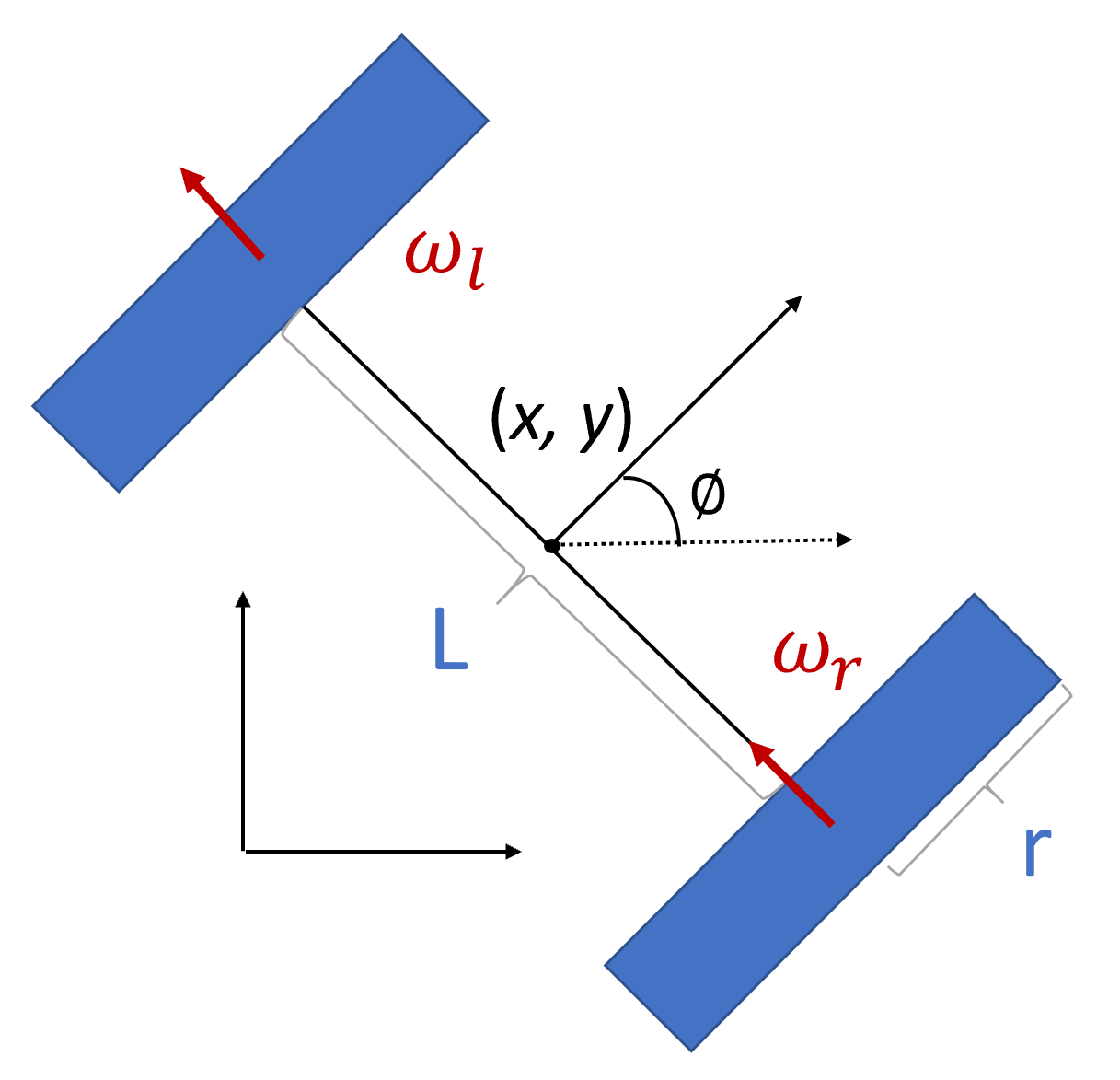}
\vspace{-6pt}
\caption{ROSbot 2.0 (left) and the differential drive model (right).}
\label{fig:rosbot}
\vspace{-12pt}
\end{figure}

The differential drive robot model is\,\footnote{The model is used as the baseline to create synthetic training data and to get the `true state' output in simulated testing and evaluation.}
\begin{equation}\label{system:diff}
\begin{cases}
\begin{array}{l}{\dot{x}=\frac{r}{2}(\omega_r+\omega_l)\cos(\phi) } \\ {\dot{y}=\frac{r}{2}(\omega_r+\omega_l)\sin(\phi) }\\{\dot{\phi}=\frac{r}{L}(\omega_r-\omega_l)}
\end{array}
\end{cases}
\end{equation}
where parameters $r = 0.062$\;m and $L = 0.228$\;m match those of the physical ROSbot 2.0 (Fig.~\ref{fig:rosbot}). The control input is $u=[\omega_r,\omega_l]$, with $\omega_r$ and $\omega_l$ being the angular velocities of the right and left wheel, respectively. The state vector contains position $\{x,y\}$ and orientation $\phi$.

The configuration of the chassis of the differential drive robot model is the Cartesian product of two points and a circle, i.e. $\mathbb{R}^2\times \mathbb{S}^1$. We first map the orientation $\phi$ to the implicit parameterization $\{\sin\phi,\cos\phi\}$. Then, the dictionary of lifting functions is constructed as the Kronecker product of Hermite polynomials of the four variables $\{x,y,\sin\phi,\cos\phi\}$ of up to first order terms (as shown in Table~\ref{tab:configuration_space}).

For operator learning we simulate $100$ trajectories each lasting for $50$\;sec with sampling rate $T=0.1s$. To construct the training data, we input a random signal to~\eqref{system:diff} that is sampled from the normal distribution $u\sim \mathcal{N}([0,0]^T,9^2I_{2\times 2})$. The covariance magnitude is selected so as to decrease chances of generating control inputs that would be unattainable by the robot or damage it, or could lead to damage to, the physical robot. The learned operator is validated by picking random input signals of length $L=50$ sampled from the same normal distribution as in the training set. We calculate the Mean Squared Error (MSE) between predicted (superscript $p$) and true (superscript $t$) states per $MSE = \frac{1}{L}\sum_1^L ([x^{p}, y^{p},\phi^{p}]-[x^t,y^{t},\phi^{t}])^2$. The MSE of the validation set is very low, 
$MSE = $ [$0.0097$\;mm, $0.0000$\;mm, $0.0013$\;rad].

\begin{figure}[h]
\vspace{-6pt}
\centering
\includegraphics[trim={0.75cm 0.25cm 1.8cm 0cm},clip,width = 0.32\textwidth]{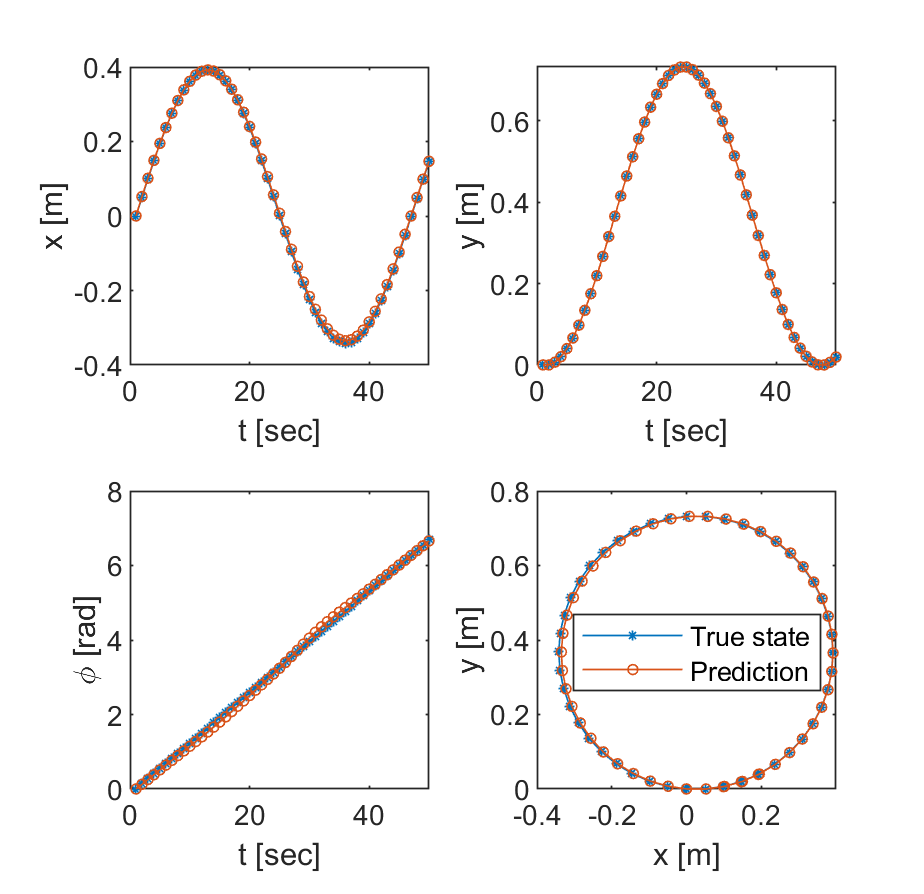}
\vspace{-12pt}
\caption{Results from testing ACD-EDMD's generalization capacity in making a simulated differential-drive robot follow a circular trajectory using random-input-signal simulated training data.}
\label{fig:dd_circle}
\vspace{-6pt}
\end{figure}

We then evaluate the learned operator's generalization capacity by testing with input $u =[12.1369,6.7310]$\;rad/sec. This input is designed to make the robot follow a counter-clockwise circular trajectory (starting from the origin). The predicted (by our method) and true (via~\eqref{system:diff}) states are shown in Fig.~\ref{fig:dd_circle}. Comparison results with other approaches are listed in Table~\ref{tab:DD_comp}. Results indicate that the model learned from the Koopman operator via ACD-EDMD is able to predict a trajectory very distinct from what it was trained on with very small error. In contrast, the direct combination of Hermite polynomials leads to large errors, while EDMD-DL and SINDy can achieve small errors (though still much larger than ACD-EDMD) but require much longer training times.

\subsection{Experiments with the Physical ROSbot 2.0 Robot}
Next, we move on with evaluating the learned operator's generalization capacity by testing using actual experimental data from the physical robot. Note that the dictionary is the same as in the simulated case above, constructed on the basis of random inputs using~\eqref{system:diff}. The training dataset is captured based on a random chattering linear velocity and angular velocity input signals for $10$ continuous trials.
We consider two evaluation cases; \textbf{Case 1}: predicting the same counter-clockwise circular trajectory as above; and \textbf{Case 2}: predicting a sharp turn trajectory. 
We wish to highlight that both trajectories are very distinct from what the learned operator has been trained upon, and contain noise (e.g., the effect of unmodeled dynamics such as friction) which is also not captured by training using the nominal model~\eqref{system:diff}. 

\begin{table}
    \caption{Comparative Results in Simulated Differential Drive Robot} \label{tab:DD_comp}
    \vspace{-6pt}
    \centering
    \begin{tabular}{l c r}
    \toprule
        \multirow{2}{*}{Method} & MSE & Training Time \\
         & [mm; mm; rad] & [sec] \\
         \midrule
         Hermite Polynomials & [291.063; 110.825; 0] & 2.94 \\
         \midrule
         EDMD-DL & [71.616; 1.119; 0.0624]& 2080.11\\
         \midrule
         SINDy & [1.495; 1.729; 0]& 2631.00\\
        \midrule 
        ACD-EDMD (ours) &[0.035; 0; 0.006]&1.97\\
        \bottomrule
    \end{tabular}
    \vspace{-9pt}
\end{table}

\begin{figure}[!h]
\vspace{0pt}
\centering
\includegraphics[trim={1cm 0.25cm 0.2cm 1cm},clip,width = 0.32\textwidth]{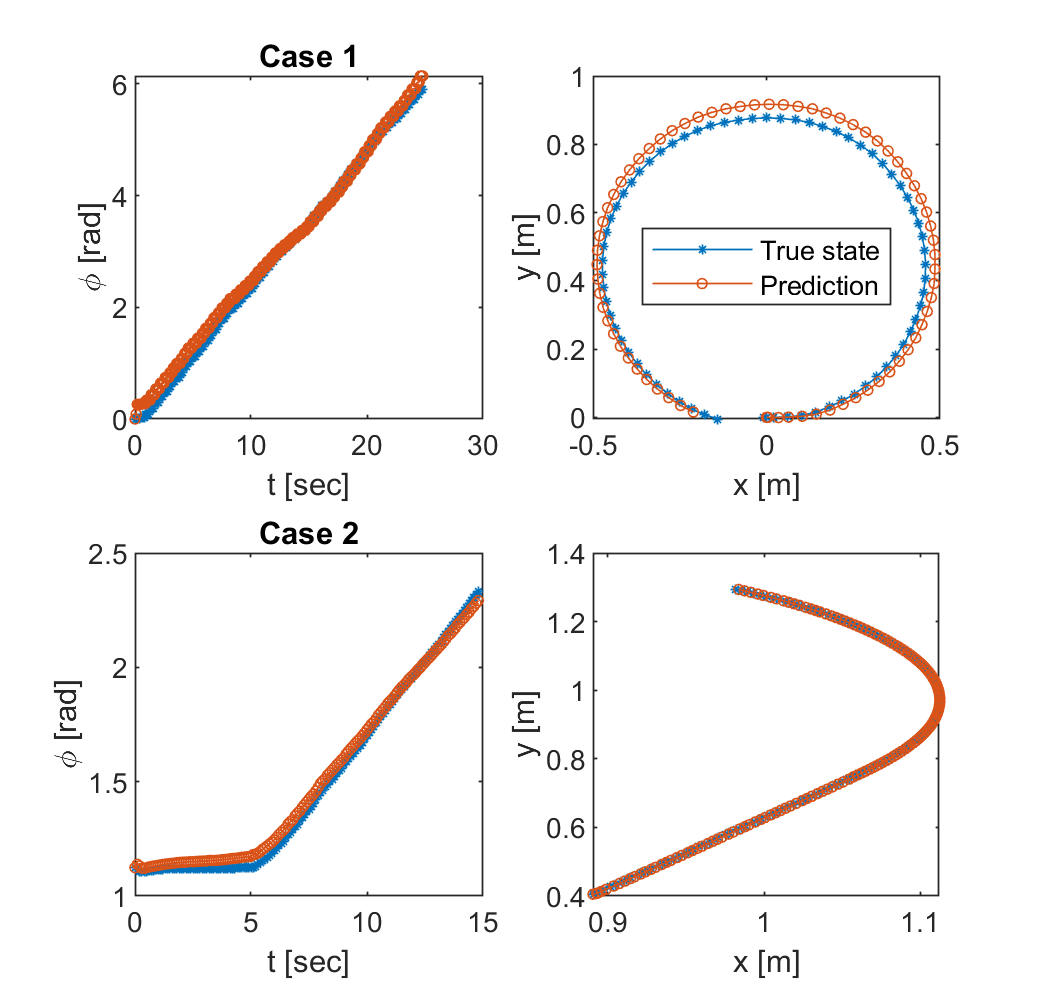}
\vspace{-12pt}
\caption{Results from testing ACD-EDMD's generalization capacity in making ROSbot follow a circular (top panels) and a sharp turn trajectory (bottom panels) using random-input-signal simulated training data.
}
\label{fig:rb}
\vspace{-6pt}
\end{figure}

The true states (obtained by remote-controlling the physical robot\,\footnote{Experimental data were collected using a motion capture camera system.}) and the predicted states (obtained by our method using logged control inputs from the experiment) are shown in Fig.~\ref{fig:rb}. 
Results indicate that the model learned from the Koopman operator using random-input simulated data is able to predict \emph{experimental trajectories}, which are also very distinct from the training dataset, with very small error (Case 1 $MSE = $ [$0.0008$\;m, $0.0011$\;m, $0.0211$\;rad]; Case 2 $MSE = $ [$0.0016$\;mm, $0.0154$\;mm, $0.0011$\;rad]). There exist some small discrepancies in the circular trajectory (Case 1, top panels in Fig.~\ref{fig:rb}), but overall results predicted using our proposed method capture accurately experimentally the  observed trajectories.

\subsection{Two Revolute Joint (2R) Rigid Robotic Arm - Simulation}
We further evaluate our method by testing with a topologically distinct from the wheeled robot system; a 2R rigid robotic arm. 
The input signal contains the joint torques, that is $\tau = [\tau_1,\tau_2]$. The state vector contains the joint angles $\theta= [\theta_1, \theta_2]$ and their first derivatives (joint angular velocities) $\dot\theta=[\dot{\theta}_1$ $\dot{\theta}_2]$. The Lagrangian method~\cite{lynch2017MR} yields 
$M(\theta) =$
\begin{equation*}
    \begin{medsize}
        \left[ \begin{matrix} m_1L_1^2+m_2(L_1^2+2L_1L_2\cos\theta_2+L_2^2)&m_2(L_1L_2\cos\theta_2+L_2^2) \\ m_2(L_1L_2\cos\theta_2+L_2^2) &m_2L_2^2 \end{matrix}
        \right],
    \end{medsize}
\end{equation*}
and
\begin{equation*}
        c(\theta,\dot{\theta})=\left[ \begin{matrix} -m_2L_1L_2\sin\theta_2(2\dot{\theta}_1\dot{\theta}_2+\dot{\theta}_2^2) \\ m_2L_1L_2\dot{\theta}_1^2\sin\theta_2 \end{matrix}
        \right],
\end{equation*}
\begin{equation*}
        g(\theta)=\left[ \begin{matrix} (m_1+m_2)L_1g\cos\theta_1+m_2gL_2\cos(\theta_1+\theta_2) \\ m_2gL_2\cos(\theta_1+\theta_2) \end{matrix}
        \right],
\end{equation*}
then the forward dynamics is given by
\begin{equation}\label{system:arm}
    \ddot{\theta} = M^{-1}(\theta)(\tau-c(\theta,\dot{\theta})-g(\theta))\enspace.
\end{equation}

The configuration space of the 2R arm is the 2-D torus and is homeomorphic to the Cartesian product of two circles, i.e. $\mathbb{T}^2 = \mathbb{S}^1\times \mathbb{S}^1$. We first express the two joint angles as the unit complex number, that is $\theta_1\mapsto\{\sin\theta_1, \cos\theta_1\}$, and $\theta_2\mapsto\{\sin\theta_2, \cos\theta_2\}$. 
Then, the dictionary of lifting functions is constructed as the Kronecker product of Hermite polynomials of the four variables $\{\sin\theta_1, \cos\theta_1, \sin\theta_2, \cos\theta_2\}$ of up to first order terms (as shown in Table~\ref{tab:configuration_space}).

For operator learning we simulate $100$ trajectories each lasting for $2$\;sec with sampling rate $T_a=0.01$\;sec; thus, each trajectory time series has length $L_{a}=200$. To construct the training data, we input a random signal to~\eqref{system:arm}, sampled from the standard normal distribution (i.e. $\tau \sim \mathcal{N}([0,0]^T,I_{2\times2})$).

The learned operator is validated by picking random input (time series) signals of length $L_a=100$ sampled from the standard normal distribution as in the training set. We calculate the MSE between predicted (superscript $p$) and true (superscript $t$) states per $MSE = \frac{1}{L_a}\sum_{1}^{L_a} ([\theta_1^{p},\theta_2^{p}]-[\theta_1^{t},\theta_2^{t}])^2$). The MSE of the validation set is very low, $MSE = $ [$0.0003$\;rad, $0.0009$\;rad]. 

\begin{figure}[h]
\vspace{2pt}
\centering
\includegraphics[trim={0.25cm 0.1cm 0.25cm 0.1cm},clip,width = 0.175\textwidth]{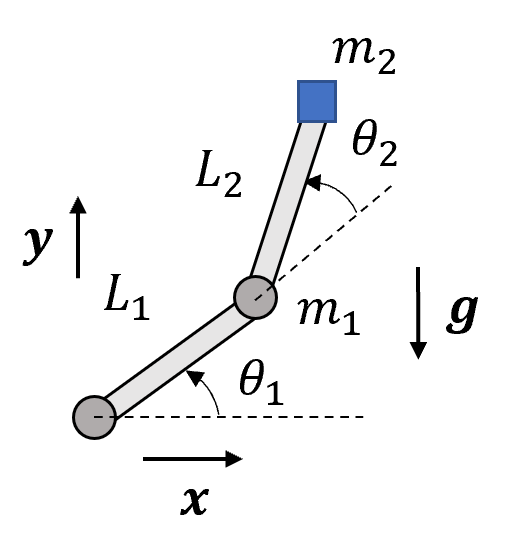}\includegraphics[trim={0.25cm 0.1cm 0.25cm 0.1cm},clip,width = 0.2\textwidth]{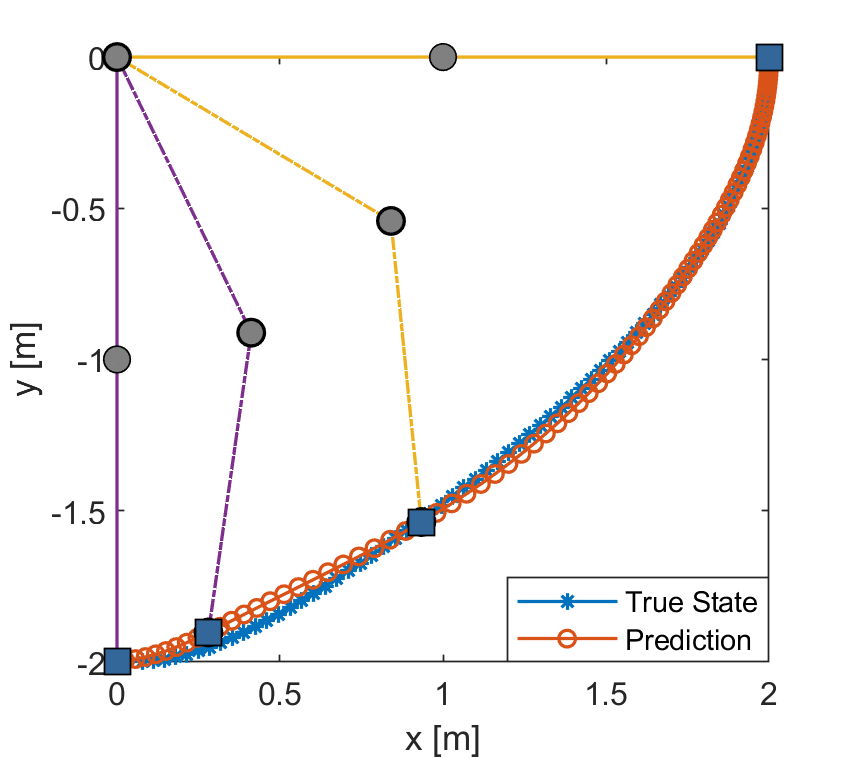}
\vspace{-9pt}
\caption{Model of 2R robotic arm (left panel). Results from testing generalization capacity in driving the arm to follow a clockwise quadrant trajectory (right panel). The yellow and purple solid lines indicate the initial and final positions. The dashed lines indicate intermediate states of the arm.
}
\label{fig:r2_quadrant}
\end{figure}

We then evaluate the learned operator's generalization capacity by testing with input $\tau =[-1,0]$\;Nm. This input is designed to make the arm's end-effector follow a clockwise circular trajectory (from a horizontal to a vertical configuration in the fourth quadrant). The predicted (by our method) and true (via~\eqref{system:arm}) states are shown in Fig.~\ref{fig:r2_quadrant}. Comparison results are illustrated in Table~\ref{tab:arm_comp}. 
Results indicate that the model learned via ACD-EDMD is able to predict a trajectory very distinct from what it was trained on with very small error. While EDMD with Hermite polynomials requires the shortest time (approximately half of ACD-EDMD but same order of magnitude) it leads to larger (over one order of magnitude) errors in $\theta_2$. In contrast, EDMD-DL and SINDy produce smaller errors but at the expense of increasing the training time for more than two orders of magnitude.

\begin{table}[]
\vspace{-6pt}
\caption{Comparative Results in Simulated 2R Arm} \label{tab:arm_comp}
    \vspace{-6pt}
    \centering
    \begin{tabular}{l c r}
    \toprule
        \multirow{2}{*}{Method} & MSE & Training Time \\
         & [rad; rad] & [sec] \\
         \midrule
         Hermite Polynomials &[0.0016; 0.0594] & 0.59\\
         \midrule
         EDMD-DL & [0.0000; 0.0000]& 280.45\\
         \midrule
         SINDy& [0.0000; 0.0016] & 576.92\\
         \midrule
         ACD-EDMD (ours) &[0.0015; 0.0036]& 0.96\\
         \bottomrule
    \end{tabular}
    \vspace{-12pt}
\end{table}


\subsection{Experiments with a Soft Robotic Leg}
Lastly yet importantly, we turn our attention to soft robotic systems. In contrast to rigid robotic systems whereby their configuration space can be uniquely determined, soft robotic systems do not possess such property. However, the workspace of a soft robot can be determined uniquely and hence we propose employing the workspace topology in the case of soft robotic systems.
\begin{figure}[h]
\vspace{-6pt}
\centering
\includegraphics[trim={0cm 0.25cm 0cm 0.2cm},clip,width = 0.225\textwidth]{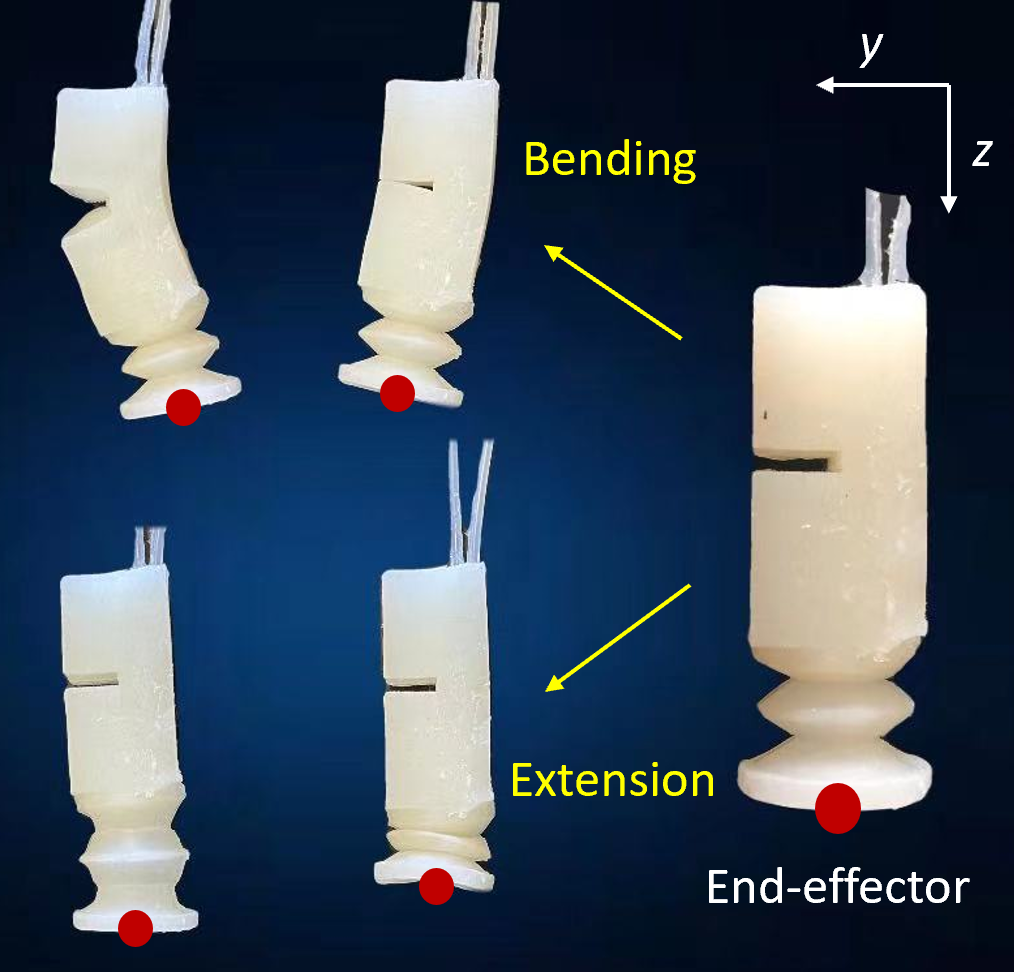}
\vspace{-6pt}
\caption{The soft robotic leg considered herein shown at distinct operating settings (left: pressurization; middle: depressurization; right: idle).}
\label{fig:overview_leg}
\vspace{-6pt}
\end{figure}

The soft robotic system we consider is a soft pneumatic robotic leg~\cite{liu2020SoRX}. The leg comprises of 1) the bending part, and 2) the extension part (Fig.~\ref{fig:overview_leg}). When the two parts are simultaneously pressurized, the leg both bends and extends. Pressurization is controlled by two DC 12V 2-way normally closed electric solenoid air valves, one for bending ($|u_1|$) and one for extension ($|u_2|$). Two vacuum pumps are utilized for input (+) and exhaust (-). A motion capture camera system is used to collect position data $(y,z)$ of the tip of the soft leg. The workspace of the tip is $\mathbb{R}^2$ and we observe both $y$ and $z$. Hence, the dictionary to be used in ACD-EDMD is formed as the Kronecker product of $H_1(y)$ and $H_1(z)$.

We adopt a two-pronged training and evaluation method. First, we conduct individual tests for the two parts by setting one input at zero. Then, we pressurize/depressurize both bending and extension parts to mimic a foot path~\cite{liu2020SoRX}. 
In training, a constant voltage input signal is given to the valves that control active leg parts. Each signal lasts for $2$\;sec; training inputs for different tests are listed in Table~\ref{tab:experiment}. Note that in Case 3 we train the system with both positive and negative listed values. Collected data are post-processed by a moving-average filter with window length of $5$. 

   \begin{table}[!h]
   \vspace{-6pt}
        \centering
        \caption{Training Inputs for the Soft Robotic Leg Experiments}
        \label{tab:experiment}
        \vspace{-6pt}
        \begin{tabular}{l l}
            \toprule
            Case 1: Individual test & $u_1=+[3.12,3.61,3.64,5.92,7.97]V$\\
            for bending part&$u_2 = [0.00,0.00,0.00,0.00,0.00]V$\\
            \midrule
            Case 2: Individual test  &$u_1=[0.00,0.00,0.00,0.00,0.00]V$\\
            for extension part& $u_2 = +[3.12,3.64,4.30,5.92,6.92]V$\\
            \midrule
            Case 3: Test on the& $|u|=[(3.85,4.55),(3.85,12.03),$\\
            the whole robot leg & $(4.63,4.63),(6.16,6.16),(9.30,9.30),$\\
           &$(11.24,4.55),(11.24,12.03)]V$\\
            \bottomrule
        \end{tabular}
        \vspace{-6pt}
    \end{table}

We evaluate ACD-EDMD's generalization capacity in three cases. In Cases 1 and 2, the input signal is a combination of two constant voltages (not used in training) each lasting for $1$\;sec. 
Case 1 testing input (bending part) is $u_1 = 4.30$\;V for $1$\;sec followed by $u_1 = 3.19$\;V for another $1$\;sec, while $u_2 = 0$\;V for $2$\;sec.
In Case 2 (extension part) the input sequence is $u_1 = 0$\;V for $2$\;sec, while $u_2 = 3.22$\;V for $1$\;sec and $u_2 = 3.61$\;V for the following $1$\;sec. 

\begin{figure}[h]
\vspace{-9pt}
\centering
\includegraphics[clip,width = 0.35\textwidth]{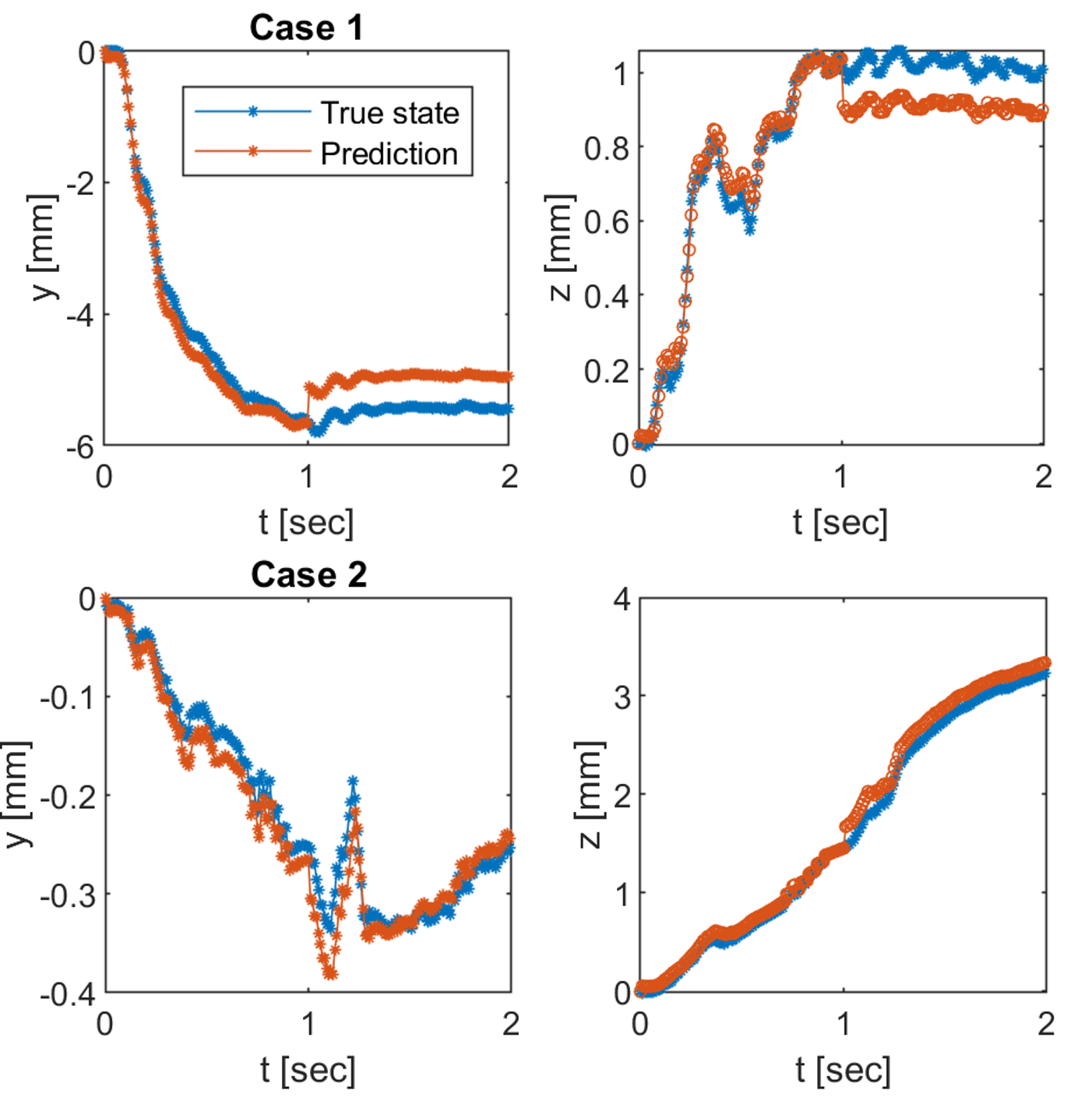}
\vspace{-12pt}
\caption{Results from testing generalization capacity when actuating only the bending part (top panels) and only the extension part (bottom panels).}
\label{fig:soft_leg}
\vspace{-6pt}
\end{figure}

\begin{figure}[!h]
\vspace{6pt}
\centering
\includegraphics[clip,width = 0.35\textwidth]{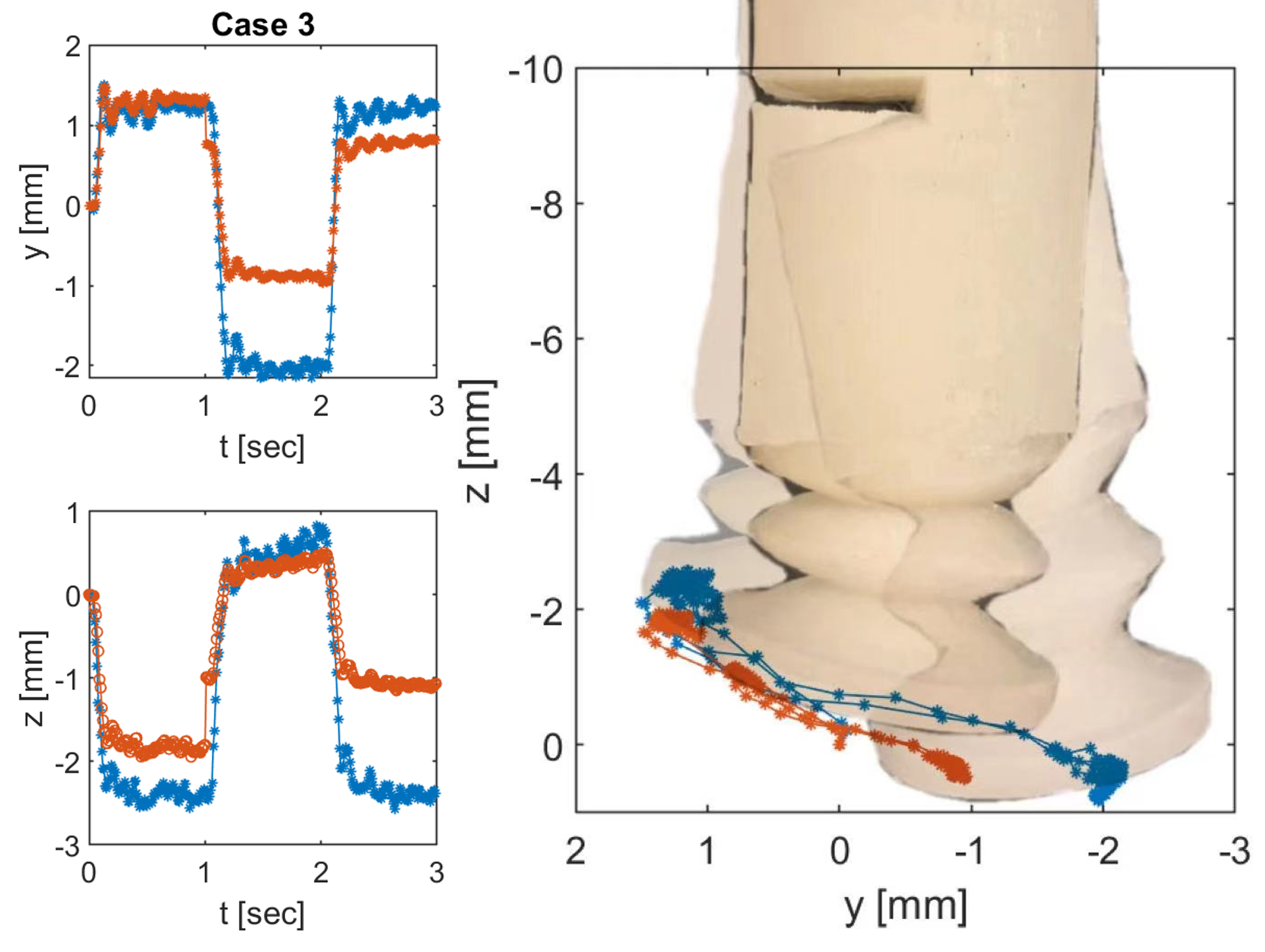}
\vspace{-12pt}
\caption{Test results for Case 3 that the robotic leg follows a walk trajectory. The blue curve with star indicates the true state and the red curve with circle represents the prediction. In the right panel we show how the leg moves; less transparent snapshots indicate later states. (Best viewed in color.)}
\label{fig:physical_robot}
\vspace{-15pt}
\end{figure}

Results in Fig.~\ref{fig:soft_leg} suggest that ACD-EDMD (in spite of using a very small dataset) offers a good prediction in the individual parts tests (Case 1 $MSE =$ [$0.1423,0.0068$]\:mm and Case 2 $MSE =$ [$0.0005, 0.0122$]\;mm). In Case 3 where both the bending and extension parts are actuated, we first drive the robot to curl up ($u = [-3.88,-7.74]$\;V) then extend as well bend ($u = [4.63,3.50]$\;V), and finally curl up again ($u = [-4.54,-3.88]$\;V) to complete one foot trajectory cycle. Results shown in Fig.~\ref{fig:physical_robot} demonstrate that the proposed method, despite its reduced performance compared to the single-part tests in Cases 1 and 2, can still predict a relatively accurate trajectory for the full actuator. 
We anticipate that a more rich training dataset can help improve prediction accuracy; improving upon these results is part of future work. 

In Table~\ref{tab:soft_comp}, we report comparative results in Case 3 between ACD-EDMD and the other related approaches. Results show that our approach achieves similar error as SINDy, but at significantly lower (two orders of magnitude) training time. EDMD-DL attains the smallest error but the training time is four orders of magnitude larger compared to ACD-EDMD's training time. Direct use of Hermite polynomials in EDMD results to the lowest accuracy. 

\begin{table}[!h]
    \vspace{-4pt}
\caption{Comparative Results in Soft Robotic Leg Experiments} 
    \label{tab:soft_comp}
    \vspace{-5pt}
    \centering
    \begin{tabular}{l c r}
    \toprule
         Method & MSE [mm; mm] & Training Time [sec]   \\
         \midrule
         Hermite Polynomials&[96.357; 2.940] & 0.11\\
         \midrule
         EDMD-DL & [0.095; 0.013]& 151.19\\
         \midrule
         SINDy & [0.507; 0.216]&2.77\\
         \midrule
         ACD-EDMD (ours) &[0.464; 0.616]&0.08\\
         \bottomrule
    \end{tabular}
    \vspace{-9pt}
\end{table}


\section{Conclusions}
The main scientific premise in this paper is that robotic systems exhibit certain characteristic properties which can be exploited to inform selection and design of appropriate lifting functions for use in the context of modeling and control of data-driven Koopman operator based robotic systems. 
Selection of an appropriate set of lifting functions directly affects the Koopman operator's ability to extract dynamics from data, and to-date several successful approaches using Koopman operator theory in the context of robotics employ empirically-selected lifting functions.

In this paper we provide an analytical way to design lifting functions based on fundamental topological spaces for robots (i.e. their configuration space if rigid and their workspace if soft) using products of Hermite polynomials. Our proposed method, termed ACD-EDMD, is simple to implement and enjoys provable guarantees of completeness and convergence when the observables are weighted bounded
Evaluation results using a range of diverse robotic systems and tested both in simulation and via physical hardware experiments reveal that our proposed method can generalize well and predict accurately the robot's state evolution. This finding is persistently observed especially in rigid robots, and even when testing the method's generalization capacity in very different cases. An example is the wheeled robot learning to follow specific patterns (e.g., a circular trajectory) while trained on (simulated) random-input-signal trajectories. 

Comparison with other related dynamics identification methods indicates that ACD-EDMD can achieve high prediction accuracy and low training times in all tested cases. Prediction accuracy is overall much higher than the baseline method of using directly EDMD with a sum of Hermite polynomials, on par to SINDy and in some cases higher than EDMD-DL (although the latter requires manual tuning of hyper-parameters to perform well). 
Training time is overall on par to the baseline case and at least two orders of magnitude faster than SINDy and two-four orders of magnitude faster than EDMD-DL. The fast training times of ACD-EDMD, compared with its high accuracy, make it beneficial for implementation in physical robots and toward online Koopman-based modeling and control architectures.

Promising results obtained herein lay the foundations to study higher-dimensional problems in future work. We demonstrated that Kronecker products of Hermite polynomials of up to first order terms can work well in the planar problems considered herein. Future work will investigate the trade-offs of including higher-order Hermite polynomial terms for prediction in 2.5D and 3D robotics problems.

\bibliographystyle{IEEEtran}
\bibliography{IEEEabrv,ral21_lskk}

\begin{thebibliography}{10}
\providecommand{\url}[1]{#1}
\csname url@rmstyle\endcsname
\providecommand{\newblock}{\relax}
\providecommand{\bibinfo}[2]{#2}
\providecommand\BIBentrySTDinterwordspacing{\spaceskip=0pt\relax}
\providecommand\BIBentryALTinterwordstretchfactor{4}
\providecommand\BIBentryALTinterwordspacing{\spaceskip=\fontdimen2\font plus
\BIBentryALTinterwordstretchfactor\fontdimen3\font minus
  \fontdimen4\font\relax}
\providecommand\BIBforeignlanguage[2]{{%
\expandafter\ifx\csname l@#1\endcsname\relax
\typeout{** WARNING: IEEEtran.bst: No hyphenation pattern has been}%
\typeout{** loaded for the language `#1'. Using the pattern for}%
\typeout{** the default language instead.}%
\else
\language=\csname l@#1\endcsname
\fi
#2}}

\bibitem{kaiser2020KoopApp}
E.~Kaiser, J.~N. Kutz, and S.~L. Brunton, ``Data-driven approximations of
  dynamical systems operators for control,'' in \emph{The Koopman Operator in
  Systems and Control}.\hskip 1em plus 0.5em minus 0.4em\relax Springer, 2020,
  pp. 197--234.

\bibitem{korda2018KoopApp}
M.~Korda and I.~Mezi{\'c}, ``Linear predictors for nonlinear dynamical systems:
  Koopman operator meets model predictive control,'' \emph{Automatica},
  vol.~93, pp. 149--160, 2018.

\bibitem{abraham2019KoopApp}
I.~Abraham and T.~D. Murphey, ``Active learning of dynamics for data-driven
  control using koopman operators,'' \emph{IEEE Transactions on Robotics},
  vol.~35, no.~5, pp. 1071--1083, 2019.

\bibitem{huang2020KoopmanApp}
B.~Huang, X.~Ma, and U.~Vaidya, ``Data-driven nonlinear stabilization using
  koopman operator,'' in \emph{The Koopman Operator in Systems and
  Control}.\hskip 1em plus 0.5em minus 0.4em\relax Springer, 2020, pp.
  313--334.

\bibitem{mamakoukas2021Taylor}
G.~Mamakoukas, M.~L. Castano, X.~Tan, and T.~D. Murphey, ``Derivative-based
  koopman operators for real-time control of robotic systems,'' \emph{IEEE
  Transactions on Robotics}, 2021.

\bibitem{shi2020ccta}
L.~Shi, H.~Teng, X.~Kan, and K.~Karydis, ``A data-driven hierarchical control
  structure for systems with uncertainty,'' in \emph{IEEE Conference on Control
  Technology and Applications (CCTA)}, 2020, pp. 57--63.

\bibitem{abraham2017RKexample}
I.~Abraham, G.~De~La~Torre, and T.~D. Murphey, ``Model-based control using
  koopman operators,'' in \emph{Robotics: Science and Systems, RSS}.\hskip 1em
  plus 0.5em minus 0.4em\relax MIT Press Journals, 2017.

\bibitem{broad2020RKexample}
A.~Broad, I.~Abraham, T.~Murphey, and B.~Argall, ``Data-driven koopman
  operators for model-based shared control of human--machine systems,''
  \emph{The International Journal of Robotics Research}, vol.~39, no.~9, pp.
  1178--1195, 2020.

\bibitem{bruder2019koopsoft}
D.~Bruder, X.~Fu, R.~B. Gillespie, C.~D. Remy, and R.~Vasudevan, ``Data-driven
  control of soft robots using koopman operator theory,'' \emph{IEEE
  Transactions on Robotics}, vol.~37, no.~3, pp. 948--961, 2021.

\bibitem{bruder2019koopsoftD}
D.~Bruder, C.~D. Remy, and R.~Vasudevan, ``Nonlinear system identification of
  soft robot dynamics using koopman operator theory,'' in \emph{IEEE
  International Conference on Robotics and Automation (ICRA)}, 2019, pp.
  6244--6250.

\bibitem{bruder2020koopsoft}
D.~Bruder, X.~Fu, R.~B. Gillespie, C.~D. Remy, and R.~Vasudevan,
  ``Koopman-based control of a soft continuum manipulator under variable
  loading conditions,'' \emph{IEEE Robotics and Automation Letters}, vol.~6,
  no.~4, pp. 6852--6859, 2021.

\bibitem{haggerty2020koopsoft}
D.~A. Haggerty, M.~J. Banks, P.~C. Curtis, I.~Mezi{\'c}, and E.~W. Hawkes,
  ``Modeling, reduction, and control of a helically actuated inertial soft
  robotic arm via the koopman operator,'' \emph{arXiv preprint
  arXiv:2011.07939}, 2020.

\bibitem{castano2020koopsoft}
M.~L. Casta{\~n}o, A.~Hess, G.~Mamakoukas, T.~Gao, T.~Murphey, and X.~Tan,
  ``Control-oriented modeling of soft robotic swimmer with koopman operators,''
  in \emph{IEEE/ASME International Conference on Advanced Intelligent
  Mechatronics (AIM)}, 2020, pp. 1679--1685.

\bibitem{williams2015EDMD}
M.~O. Williams, I.~G. Kevrekidis, and C.~W. Rowley, ``A data--driven
  approximation of the koopman operator: Extending dynamic mode
  decomposition,'' \emph{Journal of Nonlinear Science}, vol.~25, no.~6, pp.
  1307--1346, 2015.

\bibitem{boyd2001polynomials}
J.~Boyd, \emph{Chebyshev and Fourier spectral methods}.\hskip 1em plus 0.5em
  minus 0.4em\relax \hspace{-3pt}Courier Corp, 2001.

\bibitem{wendland1999RBF}
H.~Wendland, ``Meshless galerkin methods using radial basis functions,''
  \emph{Mathematics of computation}, vol.~68, no. 228, pp. 1521--1531, 1999.

\bibitem{li2017ML}
Q.~Li, F.~Dietrich, E.~M. Bollt, and I.~G. Kevrekidis, ``Extended dynamic mode
  decomposition with dictionary learning: A data-driven adaptive spectral
  decomposition of the koopman operator,'' \emph{Chaos: An Interdisciplinary
  Journal of Nonlinear Science}, vol.~27, no.~10, p. 103111, 2017.

\bibitem{yeung2019learning}
E.~Yeung, S.~Kundu, and N.~Hodas, ``Learning deep neural network
  representations for koopman operators of nonlinear dynamical systems,'' in
  \emph{American Control Conference (ACC)}, 2019, pp. 4832--4839.

\bibitem{takeishi2017learning}
N.~Takeishi, Y.~Kawahara, and T.~Yairi, ``Learning koopman invariant subspaces
  for dynamic mode decomposition,'' in \emph{Proceedings of the 31st
  International Conference on Neural Information Processing Systems}, 2017, pp.
  1130--1140.

\bibitem{netto2020LieD}
M.~Netto, Y.~Susuki, V.~Krishnan, and Y.~Zhang, ``On analytical construction of
  observable functions in extended dynamic mode decomposition for nonlinear
  estimation and prediction,'' in \emph{American Control Conference
  (ACC)}.\hskip 1em plus 0.5em minus 0.4em\relax IEEE, 2021, pp. 4190--4195.

\bibitem{bollt2019geometric}
E.~M. Bollt, ``Geometric considerations of a good dictionary for koopman
  analysis of dynamical systems: Cardinality,“primary eigenfunction,” and
  efficient representation,'' \emph{Communications in Nonlinear Science and
  Numerical Simulation}, vol. 100, p. 105833, 2021.

\bibitem{karydis2015IJRR}
K.~Karydis, I.~Poulakakis, J.~Sun, and H.~G. Tanner, ``Probabilistically valid
  stochastic extensions of deterministic models for systems with uncertainty,''
  \emph{The International Journal of Robotics Research}, vol.~34, no.~10, pp.
  1278--1295, 2015.

\bibitem{karydis2016ISER}
K.~Karydis and M.~A. Hsieh, ``Uncertainty quantification for small robots using
  principal orthogonal decomposition,'' in \emph{International Symposium on
  Experimental Robotics}.\hskip 1em plus 0.5em minus 0.4em\relax Springer,
  2016, pp. 33--42.

\bibitem{koopman1931koopman}
B.~O. Koopman, ``Hamiltonian systems and transformation in hilbert space,''
  \emph{Proceedings of the national academy of sciences of the united states of
  america}, vol.~17, no.~5, p. 315, 1931.

\bibitem{proctor2018EDMDc}
J.~L. Proctor, S.~L. Brunton, and J.~N. Kutz, ``Generalizing koopman theory to
  allow for inputs and control,'' \emph{SIAM Journal on Applied Dynamical
  Systems}, vol.~17, no.~1, pp. 909--930, 2018.

\bibitem{li2017EDMDLearning}
Q.~Li, F.~Dietrich, E.~M. Bollt, and I.~G. Kevrekidis, ``Extended dynamic mode
  decomposition with dictionary learning: A data-driven adaptive spectral
  decomposition of the koopman operator,'' \emph{Chaos: An Interdisciplinary
  Journal of Nonlinear Science}, vol.~27, no.~10, p. 103111, 2017.

\bibitem{brunton2016SINDy}
S.~L. Brunton, J.~L. Proctor, and J.~N. Kutz, ``Discovering governing equations
  from data by sparse identification of nonlinear dynamical systems,''
  \emph{Proceedings of the national academy of sciences}, vol. 113, no.~15, pp.
  3932--3937, 2016.

\bibitem{brunton2016SINDyc}
------, ``Sparse identification of nonlinear dynamics with control (sindyc),''
  \emph{IFAC-PapersOnLine}, vol.~49, no.~18, pp. 710--715, 2016.

\bibitem{klus2016convergence}
S.~Klus and C.~Sch{\"u}tte, ``Towards tensor-based methods for the numerical
  approximation of the perron-frobenius and koopman operator,'' \emph{Journal
  of Computational Dynamics}, 2016.

\bibitem{korda2018convergence}
M.~Korda and I.~Mezi{\'c}, ``On convergence of extended dynamic mode
  decomposition to the koopman operator,'' \emph{Journal of Nonlinear Science},
  vol.~28, no.~2, pp. 687--710, 2018.

\bibitem{celeghini2021hermite}
E.~Celeghini, M.~Gadella, and M.~A. del Olmo, ``Hermite functions and fourier
  series,'' \emph{Symmetry}, vol.~13, no.~5, p. 853, 2021.

\bibitem{reed1972hermite}
B.~Reed, M.;~Simon, ``Functional analysis,'' \emph{Academic Press:}, 1972.

\bibitem{jing2017overview}
Z.~Jing, L.~Qiao, H.~Pan, Y.~Yang, and W.~Chen, ``An overview of the
  configuration and manipulation of soft robotics for on-orbit servicing,''
  \emph{Science China Information Sciences}, vol.~60, no.~5, p. 050201, 2017.

\bibitem{maheshwari2014Hermite}
P.~Maheshwari, G.~Mukhopadhyay, and S.~SenGupta, ``Properties of tensor hermite
  polynomials,'' \emph{arXiv preprint arXiv:1411.7398}, 2014.

\bibitem{liu2020SoRX}
Z.~Liu, Z.~Lu, and K.~Karydis, ``Sorx: A soft pneumatic hexapedal robot to
  traverse rough, steep, and unstable terrain,'' in \emph{IEEE International
  Conference on Robotics and Automation (ICRA)}, 2020, pp. 420--426.

\bibitem{tibshirani1996LASSO}
R.~Tibshirani, ``Regression shrinkage and selection via the lasso,''
  \emph{Journal of the Royal Statistical Society: Series B (Methodological)},
  vol.~58, no.~1, pp. 267--288, 1996.

\bibitem{lynch2017MR}
K.~M. Lynch and F.~C. Park, \emph{Modern Robotics}.\hskip 1em plus 0.5em minus
  0.4em\relax Cambridge University Press, 2017.

\end{thebibliography}

\end{document}